\documentclass{article}

\usepackage[preprint]{neurips_2024}

\usepackage[utf8]{inputenc} 

\usepackage{lineno,subcaption}
\usepackage[T1]{fontenc}
\usepackage{graphicx}
\usepackage{booktabs}
\usepackage[misc]{ifsym}

\usepackage{mwe}

\usepackage{url}
\usepackage{amsfonts, amsmath, amssymb}
\usepackage{mathtools}
\usepackage{nicefrac}       
\usepackage{microtype}      
\usepackage{xcolor}  
\usepackage{booktabs}
\usepackage{float}
\usepackage{hyperref}

\usepackage{listings}
\usepackage{dirtree}

\definecolor{codegreen}{rgb}{0,0.6,0}
\definecolor{codegray}{rgb}{0.5,0.5,0.5}
\definecolor{codepurple}{rgb}{0.58,0,0.82}
\definecolor{backcolour}{rgb}{0.95,0.95,0.92}

\lstdefinestyle{mystyle}{
    backgroundcolor=\color{backcolour},   
    commentstyle=\color{codegreen},
    keywordstyle=\color{magenta},
    numberstyle=\tiny\color{codegray},
    stringstyle=\color{codepurple},
    basicstyle=\ttfamily\footnotesize,
    breakatwhitespace=false,         
    breaklines=true,                 
    captionpos=b,                    
    keepspaces=true,                                 
    showspaces=false,                
    showstringspaces=false,
    showtabs=false,                  
    tabsize=2
}

\newcommand{\revision}[1]{\textcolor{black}{#1}}
\newcommand{\F}{\mathcal{F}}

\newcommand{\D}{\mathcal{D}}

\newcommand{\R}{\mathbb{R}}

\newcommand{\cov}{\mathrm{Cov}}

\newcommand{\x}{\mathbf{x}}
\newcommand{\s}{\mathbf{s}}

\newcommand{\MA}{\mathrm{MA}}

\DeclareMathOperator{\bias}{bias}

\newcommand{\indep}{\perp \!\!\! \perp}
\newcommand{\dep}{\not \! \indep}

\newtheorem{definition}{Definition}[section]

\title{GECOBench: A Gender-Controlled Text Dataset and Benchmark for Quantifying Biases in Explanations}

\author{%
  Rick Wilming \\
  Physikalisch-Technische Bundesanstalt\\
  Abbestr. 2–12, 10587 Berlin, Germany\\
  \And
  Artur Dox \\
  Physikalisch-Technische Bundesanstalt\\
  Abbestr. 2–12, 10587 Berlin, Germany\\
  \AND
  Hjalmar Schulze \\
    Physikalisch-Technische Bundesanstalt\\
  Abbestr. 2–12, 10587 Berlin, Germany\\
  \And
  Marta Olivaira \\
    Physikalisch-Technische Bundesanstalt\\
  Abbestr. 2–12, 10587 Berlin, Germany\\
  \And
  Benedict Clark \\
    Physikalisch-Technische Bundesanstalt\\
  Abbestr. 2–12, 10587 Berlin, Germany\\
  \And
  Stefan Haufe \\
    Physikalisch-Technische Bundesanstalt\\
  Abbestr. 2–12, 10587 Berlin, Germany\\
  Technische Universität Berlin\\
  Str. des 17. Juni 135, 10623 Berlin, Germany
}

\begin{document}

\maketitle

\begin{abstract}

Large pre-trained language models have become a crucial backbone for many downstream tasks in natural language processing (NLP), and while they are trained on a plethora of data containing a variety of biases, such as gender biases, it has been shown that they can also inherit such biases in their weights, potentially affecting their prediction behavior.
However, it is unclear to what extent such biases also impact \revision{feature attributions}, generated by applying ``explainable artificial intelligence'' (XAI) techniques, in possibly unfavorable ways.
To systematically study this question, we create a gender-controlled text dataset, GECO, in which the alteration of grammatical gender forms induces class-specific words and gives rise to ground truth \revision{feature attributions} for gender classification tasks, enabling the objective evaluation of the correctness of XAI methods. 
We apply this dataset to the pre-trained BERT model, which we fine-tune to different degrees, to quantitatively measure how pre-training induces undesirable bias in \revision{feature attributions} and to what extent fine-tuning can mitigate such explanation bias.
To this extent, we provide GECOBench, a rigorous quantitative evaluation framework for benchmarking popular XAI methods. 
We show a clear dependency between explanation performance and the number of fine-tuned layers, where XAI methods are observed to benefit from fine-tuning or complete retraining of embedding layers, particularly. 

\end{abstract}

\section{Introduction}
Large neural network architectures are often complex, making it difficult to understand the mechanisms by which model outputs are generated. This has led to the development of dedicated post-hoc analysis tools that are commonly referred to as ``explainable artificial intelligence'' (XAI).
In many cases, XAI methods provide so-called feature attributions, which assign an ``importance'' score to each feature of a given input \citep[e.g.][]{ribeiroWhyShouldTrust2016,lundbergUnifiedApproachInterpreting2017,sundararajanAxiomaticAttributionDeep2017}.
For the Natural Language Processing (NLP) domain, feature attribution methods in supervised learning settings are expected to highlight parts of an input text (e.g., words or sentences) that are related to the predicted target, e.g., a sentiment score.

\revision{However, it remains unclear to what extent feature attribution methods help answer specific explanation goals, such as model debugging \citep{haufe2024explainable}.
With it, questions arise about the correctness of feature attributions.
One reason why it is challenging to determine attribution methods' correctness is the tension between model-centric and data-centric explanations \citep[e.g.][]{murdochDefinitionsMethodsApplications2019,chenTrueModelTrue2020,fryerExplainingData2020,haufe2024explainable}.
In these scenarios, it is unknown how to define notions of correctness in a principled manner; thus, the extent to which feature attribution methods provide explanations that are purely model-centric or data-centric is unknown.
Empirical studies on synthetic datasets have demonstrated that numerous feature attribution methods fail to fulfill basic data-centric requirements, such as highlighting features that have a statistical association with the prediction target (also referred to as the \emph{Statistical Association Property} (SAP)) \citep[e.g.][]{wilmingScrutinizingXAIUsing2022, oliveira2023benchmark, clarkTetris2023}.
Here, we adopt this data-centric view of assessing the correctness of feature attributions and apply it to the NLP domain.}

\revision{Furthermore, within the NLP domain} attribution methods are typically applied to large pre-trained language models, which are adapted to downstream tasks through transfer learning (e.g., BERT \citep{devlinBERTPretrainingDeep2019} and its variants \citep{devlinBERTPretrainingDeep2019, liuRoBERTaRobustlyOptimized2019, radford2018improving}).
\begin{figure*}[ht]
    \centering
    \includegraphics[width=0.9\textwidth]{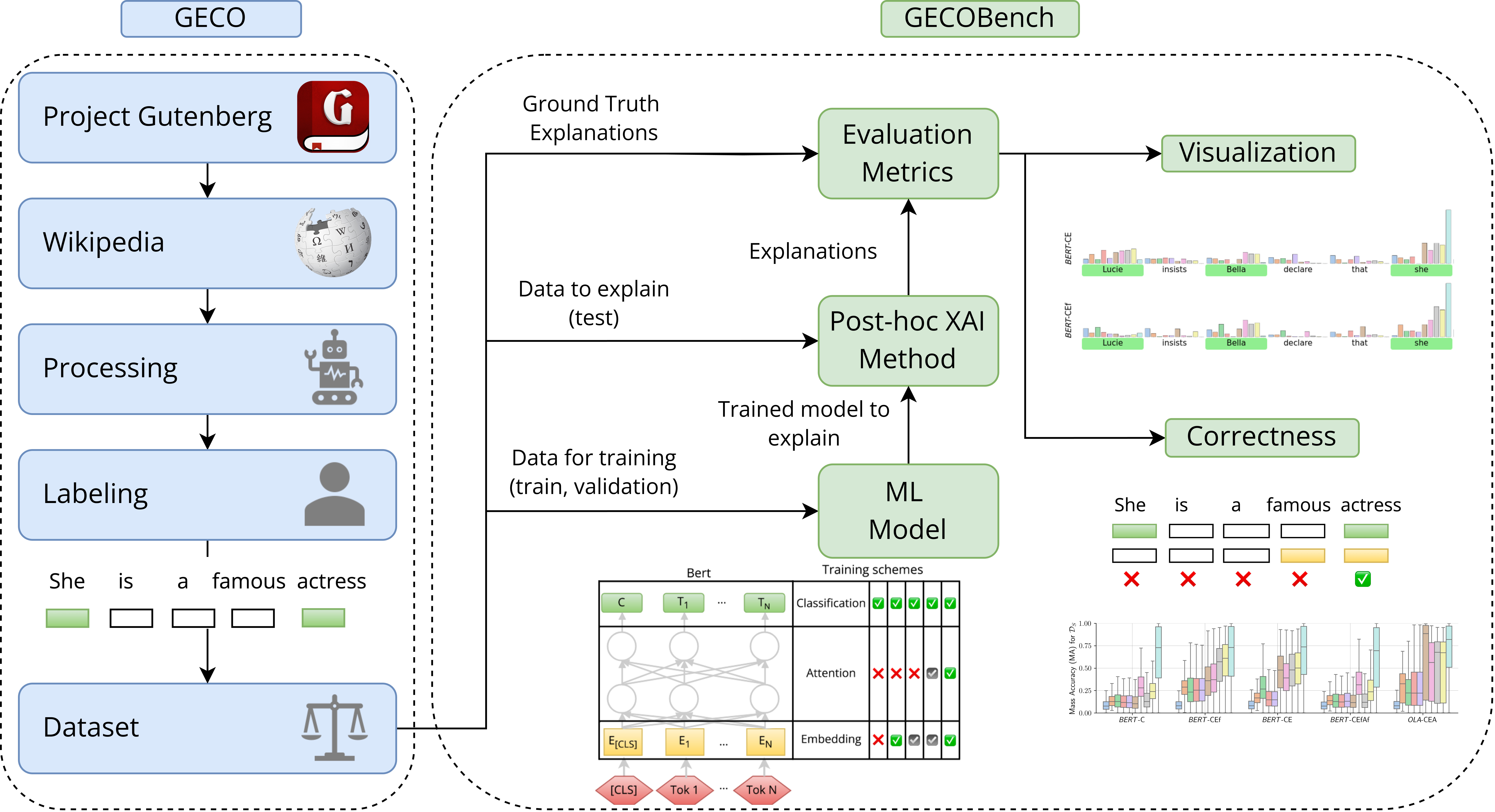}
    \caption{Overview of the benchmarking approach for evaluating the correctness of XAI methods. Starting from a clear definition of discriminative features inducing statistical associations between features/words and the prediction target, we specify ground truth explanations.
    With that, we craft a gender-focused dataset GECO, with text sourced from Wikipedia, by labeling and altering the grammatical gender of specific words.
    The resulting training and validation datasets are used to train the language model BERT. The test dataset, together with the trained model, serves as input to the XAI method, which outputs explanations for the test set. The word-based ground truth explanations, provided by the former labeling process, are then used to measure the correctness of each sentence's generated explanations using the Mass Accuracy metric \citep{arrasCLEVRXAIBenchmarkDataset2022,clarkTetris2023,clarkEXACTPlatformEmpirically2025}.
    }
    \label{fig:benchmarking-overview} 
\end{figure*}
Pre-trained language models are commonly trained on large corpora of text scraped from public and non-public sources, including Wikipedia, Project Gutenberg\footnote{https://www.gutenberg.org}, or OpenWebText\footnote{https://github.com/jcpeterson/openwebtext}.
These large corpora contain a variety of biases, such as biases against demographic groups \citep{beukeboom2014mechanisms,graells2015first,reagle2011gender}.
It has been shown that such biases affect model weights \citep{mitchell1980need,montanez2019futility} and that text corpora exhibiting problematic biases are amplified in large language models, such as BERT \citep[e.g.][]{bordiaIdentifyingReducingGender2019, gonenLipstickPigDebiasing2019,blodgettLanguageTechnologyPower2020,nadeemStereoSetMeasuringStereotypical2021}.

\revision{However, it remains unclear to what extent biases contained in pre-training corpora are reflected in explanations provided by feature attribution methods, potentially hindering them from meeting specific correctness requirements such as the SAP with respect to the target data distribution and prediction task.}
Using the example of grammatical gender, we can imagine one particular way in which pre-training biases might lead to incorrect \revision{feature attributions} or point to residual bias in fine-tuned models. 
In a gender classification task, asymmetries in the frequencies of specific words may be present in a pre-training corpus but not in the target domain. For example, historical novels may be biased towards male protagonists and depict females less frequently and in more narrowly defined roles, often adhering to historical gender norms.
However, the association between, for example, role-specific words and gender in these texts is irrelevant when it comes to distinguishing grammatical gender (as well as for many other tasks).
\revision{A feature attribution method} that highlights respective words, thus suggesting the influence of pre-training biases.

\revision{To study and quantify the data-centric correctness of feature attribution methods and the influence of biases, we make two key contributions: (1) \textit{GECO} -- a gender-controlled dataset and (2) \textit{GECOBench} -- a quantitative benchmarking framework to assess the correctness of feature attributions for language models on gender classification tasks. Both contribute to the future development of novel XAI methods, helping with their evaluation and correctness assessment. An overview is shown in Figure \ref{fig:benchmarking-overview}}.

\textit{GECO}\footnote{Available on OSF: \url{https://osf.io/74j9s/?view_only=8f80e68d2bba42258da325fa47b9010f}} is a gender-controlled dataset in which each sentence $\x$ appears in three grammatically gendered variants, male $\x^M$, female $\x^F$, and non-binary $\x^{NB}$.
The three variants are identical apart from gender-specific words such as pronouns. 
For example, consider the sentence ``She loves to spend time with her favorite cat.''
We label this sentence as ``female (`F')'' because it entails the pronouns ``she'' and ``her''.
By replacing the pronouns with ``he'' and ``his'', we define the ``male (`M')'' counterpart of this sentence.
Our approach of creating sentences with minimal changes can be seen similarly to counterfactual data augmentation \citep{kaushik2019learning,liu-etal-2021-counterfactual}.

\revision{\textit{GECOBench}\footnote{All code, including dataset generation, model training, evaluation, and visualization, is available at: \url{https://github.com/braindatalab/gecobench}} is a workflow to quantitatively benchmark the correctness of feature attributions, specifically evaluating XAI methods for NLP classification tasks induced by GECO or similar datasets.} 
Here, we are showcasing the usage of GECOBench, where BERT \citep{devlinBERTPretrainingDeep2019}, a language model pre-trained on Wikipedia data, serves as an exemplary language model.
While this benchmark can be extended to include more models, our primary focus is on benchmarking explanation methods rather than the language models themselves. 

With the gender-controlled dataset GECO, we aim to construct sentences composed of discriminative features (or words) and, therefore, ground truth \revision{feature attributions} that are gender-balanced concerning the classification task.
Further, it is known that BERT suffers from gender biases \citep{nadeemStereoSetMeasuringStereotypical2021,ahnMitigatingLanguageDependentEthnic2021}.
Thus, when using GECO as a test set, any residual asymmetry in \revision{feature attributions} can be traced back to biases induced by pre-training. 
\revision{Via the data-centric notion of correctness, we quantify this effect for different stages of re-training or fine-tuning distinct layers of BERT's architecture to investigate to what extent re-training or fine-tuning BERT on gender-controlled data can mitigate gender bias in feature attributions.}
\revision{In other words, we analyze to what extent the correctness of feature attribution is indicative of biases in models.}
By ensuring that the distinct trained models have equivalent classification accuracy throughout the considered fine-tuning stages, we can assess how these training regimes impact \revision{correctness} performance with the proposed dataset.
Generally, we do not expect any XAI method to perform perfectly, as \revision{data-centric} correctness is only one goal of interpreting machine learning models and not necessarily the primary purpose of each explanation approach. Using GECO and GECOBench, we aim to answer the following two main research questions: 
\begin{enumerate}
    \item[RQ1:] \revision{What is the explanation performance of widely adopted XAI methods in the regime of data-centric feature importance given word/token-level ground truth feature attributions?}
    \item[RQ2:] \revision{Does gender bias contained in pre-trained language models affect data-centric explanation performance of feature attribution methods, and if so, does this effect depend on the selection of layers that are fine-tuned or re-trained?}
\end{enumerate}

\section{Related Work}
Although applications of XAI have proliferated in the past years \citep[e.g.][]{lundbergExplainableMachinelearningPredictions2018,jimenezLunaDrugDiscoveryExplainable2020,tranDeepLearningCancer2021,zhangApplicationsExplainableArtificial2022}, the problems to be addressed by XAI have rarely been formally defined \citep{murdochDefinitionsMethodsApplications2019}. 
In particular, the widely-used metaphor of identifying features ``used'' by a model, measured through ``faithfulness'' or ``fidelity'' metrics \citep[e.g.][]{jacoviFaithfullyInterpretableNLP2020,hookerBenchmarkInterpretabilityMethods2019,rongConsistent2022}, can lead to fundamental misinterpretations, as such a notion depends strongly on the structure of the underlying data generative model and the resulting distribution of the (training) data \citep{haufeInterpretationWeightVectors2014,wilmingTheory2023,haufe2024explainable}. 
\citet{wilmingTheory2023} investigate such metrics, showing that many perturbation and pixel-flipping methods fail to detect statistical dependencies or other feature effects like suppressor variables \citep{friedmanGraphicalViewsSuppression2005,haufeInterpretationWeightVectors2014}, and are therefore unsuitable to directly measure certain meaningful notions of explanation ``correctness''.
To objectively evaluate whether a \revision{feature attribution} method possesses this property, the availability of ground truth data is instrumental.
Ground truth data for \revision{feature attributions} in domains such as image, tabular, and time series data have been developed in the last few years \citep[e.g.][]{kimInterpretabilityFeatureAttribution2018,ismailInputCellAttentionReduces2019,ismailBenchmarkingDeepLearning2020,tjoaQuantifyingExplainabilitySaliency2020,agarwal2022openxai,arrasCLEVRXAIBenchmarkDataset2022}. 
However, most of these benchmarks do not present realistic correlations between class-dependent and class-agnostic features (e.g., the foreground or object of an image versus the background) \citep{clarkTetris2023}, and often use surrogate metrics like faithfulness instead of directly measuring explanation performance.
Other works discuss the need for normative frameworks \citep{sullivanSides2024} or studying data manipulation and its impact on XAI methods' output \citep{mhasawadeUnderstanding2024}, rather than focusing on ground truth \revision{feature attributions}.
Several NLP-related benchmarks have been presented \citep{deyoungERASERBenchmarkEvaluate2020,rychenerQUACKIENLPClassification2020}; however, they also have certain limitations. 
In the case of \citep{deyoungERASERBenchmarkEvaluate2020}, faithfulness of the model is measured in alignment with human-annotated rationales, which do not necessarily align with statistical association -- opening the door to cognitive biases. 
\citep{rychenerQUACKIENLPClassification2020} present a benchmark dataset consisting of a question-answering task, where the ground truth \revision{feature attributions} originate from a text context providing the answer.
However, as the authors emphasize, defining a ground truth for question-answering cannot depend on only one word but rather on a context of words providing the prediction models with sufficient information. 
This work, therefore, does not provide word-level ground truth \revision{feature attributions} in the sense of statistical association.
\citep{balagopalanRoadExplainabilityPaved2022} and \citep{daiFairness2022} analyze the fairness behavior of XAI methods, focusing on model fidelity, highlighting disparities between social groups, rather than considering the correctness aspect of \revision{feature attributions}.

\revision{Moreover, \citet{joshiSaliencyGuidedDebiasing2024} propose a mitigation technique for gender bias in natural language generation based on feature attribution methods' output. 
However, since no token-level ground truth is provided, neither the correctness nor the selection-ability of biased tokens by feature attributions can be verified.}
\revision{\citet{gamboaNovelInterpretabilityMetric2024} introduce the bias attribution score, an information-theoretic metric for quantifying token-level contributions to biased behavior in multilingual pre-trained language models, demonstrating the presence of sexist and homophobic biases in these models.
Unlike GECO, neither a controlled counterfactual dataset nor ground truth attributions for evaluating the correctness of feature attribution methods are provided.}
\revision{\citet{dehdariradEvaluatingExplainabilityLanguage2025} propose a unified framework for evaluating feature attribution methods in language classification models, comparing SHAP, LIME, Integrated Gradients, and interaction-based approaches across classical and transformer architectures to assess their faithfulness \citep{samekExplainableAIInterpreting2019, jacoviFaithfullyInterpretableNLP2020} under different datasets.
However, a controlled dataset with ground truth attributions is not provided. 
Given the tension between model-centric and data-centric feature attributions, ground truth-based evaluations, as in GECOBench, allow for a more principled study of the feature attribution methods in the context of language models.}

\revision{In the NLP research community, the development of datasets for bias detection, metrics for fairness and bias assessment, and methods for bias mitigation is an active field of research. For example, \citet{bolukbasiManComputerProgrammer2016} showed that word embeddings encode gender stereotypes and proposed subspace-based debiasing, that is, learning a ``gender direction'' and projecting it out from gender-neutral words.
This was refined by \citet{prostDebiasingEmbeddingsReduced2019}.
\citet{devMeasuringMitigatingBiased2020} utilize natural language inference as a surrogate to study and mitigate biased inferences arising from embeddings systematically.
Further benchmarks for social bias-analysis in large pre-trained language models have been proposed, for example, in \citet{manziniBlackCriminalCaucasian2019, nangiaCrowSPairsChallengeDataset2020,costajussaGeBioToolkitAutomaticExtraction2020, nadeemStereoSetMeasuringStereotypical2021, parrishBBQHandbuiltBias2022, jentzschGenderBiasBERT2022,zakizadehDiFairBenchmarkDisentangled2023, navigliBiasesLargeLanguage2023, cimitanCurationBenchmarkTemplates2024}.
Yet, identical sentences for each grammatical gender, which only differ across grammatical gender forms in specific and controlled positions, providing a ground truth for feature attribution benchmarking, is not currently possible based on these datasets and benchmarks. }

\definecolor{pastelgreen}{HTML}{98daa7}
\definecolor{pastelblue}{HTML}{abc9ea}
\definecolor{pastelorange}{HTML}{efb792}
\newcommand{\changeFem}[1]{\colorbox{pastelblue}{\textbf{#1}}}
\newcommand{\changeMale}[1]{\colorbox{pastelorange}{\textbf{#1}}}
\newcommand{\changeNB}[1]{\colorbox{pastelgreen}{\textbf{#1}}}
\begin{table*}[ht]
    \centering
\begin{tabular}{llllllllll}
\toprule
\textbf{Version} & \textbf{Sentence} &  &  & & & & & &  \\
\midrule
Original  & She  & touches & the & heart & of & her & Aunt. \\
Subj. Female ($\x_S^F$)  & \changeFem{She}
  & touches & the & heart & of & \changeFem{her} & Aunt.  \\
Subj. Male ($\x_S^M$)  & \changeMale{He}  & touches & the & heart & of & \changeMale{his} & Aunt.  \\
Subj. Non-binary ($\x_S^{NB}$)  & \changeNB{They}  & \changeNB{touch} & the & heart & of & \changeNB{their} & Aunt.  \\
All Female ($\x_A^F$)  & \changeFem{She}  & touches & the & heart & of & \changeFem{her} & \changeFem{Aunt}.  \\
All Male ($\x_A^M$) & \changeMale{He}  & touches & the & heart & of & \changeMale{his} & \changeMale{Uncle}. \\
All Non-binary ($\x_A^{NB}$) & \changeNB{They}  & \changeNB{touch} & the & heart & of & \changeNB{their} & \changeNB{Parent's Sibling}. \\

\bottomrule \\
\end{tabular}
    \caption{Example of the labeling and alteration scheme of sentences, showing the original sentence and the six manipulated versions. Words marked as the ground truth for \revision{feature attributions} are written in bold and color-coded depending on the grammatical gender.} \vspace{-0.1in}
    \label{tab:examplary-sentence}
\end{table*}

\section{Methods}

To enable correctness evaluations for explanation methods, we introduce the GECO dataset, which comprises a set of manipulated sentences $\x$ in which grammatical subjects and objects assume either their male $\x^M$, female $\x^F$, or non-binary $\x^{NB}$ forms.
These three grammatically gendered variants, give rise to the downstream task of gender classification, with labels ``M'', ``F'', and ``NB'', which involves discriminating between the variants of sentences and is represented by the dataset $\D \coloneqq \{(\x^{(i,M)}, \text{``M''}), (\x^{(i,F)}, \text{``F''}), (\x^{(i, NB)}, \text{``NB''})\}_{i=1}^{n}$.
Importantly, in all cases, ground truth \revision{feature attributions} on a word-level basis are available by construction. 

\subsection{Data Sourcing \& Generation}
For the dataset, we restrict ourselves to source sentences with a human subject, such that each sentence of our manipulated dataset is guaranteed to have a well-defined gender label. This type of sentence naturally occurs in books and novels.
The Gutenberg\footnote{https://www.gutenberg.org} archive provides a plethora of classical titles that allow one to identify relevant text content from well-known novels or nonfiction titles.
To comply with licensing surrounding the listed books, we collect the content of their corresponding Wikipedia pages and only use text pieces related to the plot of the story.
We query the list of the top $100$ popular books on the Gutenberg project and obtain their corresponding Wikipedia pages. More details on data licensing are provided in Appendix \ref{app:data-licensing}.

We create two ground truth data sets $\D_S$ and $\D_{A}$. Each contains $1610$ sentences in a male, a female, and a non-binary version, comprising $4830$ sentences in total. $\D_S$ contains sentences in which \textit{only} words specifying the gender of the grammatical subject are manipulated to be either in male $\x_{S}^{M}$, female $\x_{S}^{F}$ or non-binary $\x_{S}^{NB}$ form, while $\D_A$ contains sentences in which all gender-related words are manipulated, which we denote as $\x_{A}^{M}$, $\x_{A}^{F}$ or $\x_{A}^{NB}$, respectively \revision{(see Table \ref{tab:geco-dataset})}. 
Table \ref{tab:examplary-sentence} shows an exemplary sentence and the resulting manipulations employing this labeling scheme. 
The dataset $\D_S$ instantiates a substantially more challenging task compared to dataset $\D_A$ due to the reduction of discriminative features. 
In this scenario, the model is required to differentiate between subject and object when they have different grammatical genders, necessitating a more profound understanding of the sentence’s context and structure to address the task effectively.
Thus, employing both types of datasets allows us to investigate whether the model inadvertently focuses on, for the prediction task, irrelevant parts of the sentence, potentially introducing bias that could impact explanation performance.
The process for creating these datasets consists of two consecutive steps: (i) Pre-processing of scraped Wikipedia pages.
(ii) Manual labeling is used to detect and adapt relevant subjects and objects in a sentence. 
More details on labeling and format are provided in Appendix \ref{Appendix.Gender_Dataset.Data_Generation}. Further details on hosting and future maintenance are provided in the Supplementary Material.
\begin{table}[t]
\centering
\caption{\revision{Overview of the GECO dataset properties. Each dataset variant contains sentences in three gendered forms: male (M), female (F), and non-binary (NB).}}
\label{tab:geco-dataset}
\begin{tabular}{
    p{1.0cm}  
    p{3.0cm}  
    p{1.25cm}    
    p{1.75cm}    
    p{1.1cm}  
    p{1.35cm}  
    p{1.25cm}  
}
\toprule
\textbf{Dataset} & \textbf{Manipulation Scope} & \textbf{\# Base Sent.} & \textbf{\# Variants} & \textbf{Total Sent.} & \textbf{\# Train Sent.} & \textbf{\# Test Sent.}\\
\midrule
$\D_S$ & Subject words only & $1610$ & $3$ (M, F, NB) & $4830$ & $3864$ & $966$ \\
$\D_A$ & All gender-related words & $1610$ & $3$ (M, F, NB) & $4830$ & $3864$ & $966$ \\

\bottomrule
\end{tabular}
\end{table}

\subsection{Bias Assessment of GECO}\label{Appendix.Bias_Measures}

We employ a co-occurrence metric \cite{zhaoMenAlsoShopping2017}, a rudimentary bias measure, highlighting the unbiasedness of GECO.
Specifically, we adopt the co-occurrence metric proposed by \cite{cabelloEvaluatingBiasFairness2023} to measure the gender bias of the datasets $\D_{S}$ and $\D_A$.
For a given sentence $\x \in \D$, we approximately measure the bias induced by grammatical gender when considering the co-occurrence between the sentence's gender terms and the remaining words.
First, we define a set of grammatical gender terms $A\coloneqq \{``she", ``her", ``he", ``they", \dots \}$ and second, a word vocabulary without grammatical gender terms $V \coloneqq W \setminus A$, where the vocabulary $W$ contains all words available in a corpus of $\D$, then, the co-occurrence metric $C$ is defined as
\begin{equation}
    \label{eq:co-occurrence}
    C(\D) 
    = \sum_{\x \in \D} \sum_{w \in V} \sum_{a \in A} 
    \mathbf{1}_{\x}(a, w).
\end{equation}

Further, we consider the decomposition $\D = \D^M  \cup \D^F \cup \D^{NB}$ with male $\D^M$, female $\D^F$ and non-binary $\D^{NB}$ sentences, respectively.
Then we define the bias, exemplarily for $\D^F$, according to the co-occurrence metric $C$ as
\begin{equation}
    \label{eq:co-occurrence-bias}
    \bias_C(\D^F) = \frac{C(\D^F)}{C(\D^F)+C(\D^M) + C(\D^{NB})}.
\end{equation}
A perfect balance is achieved for $\bias_C(\D^F)=1/3$, signifying that the dataset is evenly distributed between males, females, and non-binaries.
Deviations from this value indicate the presence of biases: values closer to 0 suggest a male or non-binary bias, while those approaching 1 indicate a female bias.

\subsection{Explanation Benchmarking}

The alteration of sentences induces discriminative features by construction, and their uniqueness automatically renders them the only viable ground truth \revision{feature attributions}, defining two different gender classification tasks represented by the two datasets $\D_S$ and $\D_A$. 
By extension, every word that is not grammatically gender-related, therefore not altered, becomes a non-discriminative feature.
We train machine learning models on these tasks and apply post-hoc \revision{feature attribution} methods to the trained models to obtain explanations expressing the importance of features according to each XAI method's intrinsic criteria. 
When evaluating the XAI methods, ground truth \revision{feature attributions} are adduced to measure if their output highlights the correct features. 
An overview is shown in Figure \ref{fig:benchmarking-overview}.

\subsubsection{Ground Truth \revision{Feature Attributions}}
We consider a supervised learning task, where a model $f: \R^d \to \R$ learns a function between an input $\x^{(i)} \in \R^d$ and a target $y^{(i)} \in \{-1, 1\}$, based on training data $\D = \{(\x^{(i)}, y^{(i)})\}_{i=1}^{N}$. Here, $\x^{(i)}$ and $y^{(i)}$ are realizations of the random variables $\mathbf{X}= (X_1, X_2, \dots, X_d)^{\top}$ and $Y$, with joint probability density function $p_{\mathbf{X},Y}(\x,y)$, and $[d]\coloneqq \{1, \dots, d\}$ is the set of feature indices for a vector-wise feature representation $(X_i \mid i \in [d])$. 
We formally cast the problem of finding an explanation or important features as a decision problem $([d], \F, f)$ where $\F \subseteq [d]$ is the set of \emph{important features}. 
Moreover, an explanation or saliency map $\s : \R^d \to \R^d$ should assign a numerical value reflecting the significance of each feature.
Then we are interested in finding a test $h : \R^d \to \{0, 1\}^d$, which one can use to define the set of important features $\F \coloneqq \{j \mid h_j(\x) = 1, \;\; \text{for} \;\; j \in [d]\}$.
For the concrete definition of the test and the resulting set of important features, we adopt the approach of \citep{wilmingScrutinizingXAIUsing2022,wilmingTheory2023} and give the following definition.

\begin{definition}[Statistical Association Property (SAP)]\label{def:sap}
Given the supervised learning task from above, we say that an XAI method has the Statistical Association Property (SAP) if for any feature $X_j$ with non-zero (or, significantly larger than zero) importance, there also exists a statistical dependency between $X_j$ and the target $Y$, i.e., $ X_j \dep Y$. 
\end{definition}
This definition is based on the discussion that most \revision{feature attribution} methods implicitly or explicitly assume that such a statistical association exists \citep{wilmingScrutinizingXAIUsing2022}.
Now,  defining a test via $h_j(\x) = 1$ if $X_j \dep Y$ and $h_j(\x)=0$ otherwise, we can summarize the set of \emph{potentially} important features via their univariate statistical dependence with the target $\F = \{j \mid X_j \dep Y, \,\, \text{for} \,\, j \in [d]\}$. 
Thus, each sentence of the GECO corpus and its corresponding token sequence $\x^{(i)}$ has a matching ground truth map $h(\x^{(i)}) \in \{0, 1\}^{d}$, representing corresponding important tokens.

\subsubsection{\revision{Feature Attribution} Methods}

Here, we focus on post-hoc attribution methods, which can be broadly divided into gradient-based methods and local sampling or surrogate approaches.
Generally, these methods produce an explanation $\s: \R^{d} \to \R^d$, which is a mapping that depends on the model $f$ and an instance $\x^*$ to be explained.
Gradient-based methods locally approximate a differentiable model $f$ around a given input sequence $\x^*$. 
From this class, we consider Saliency \citep{simonyan2013deep}, InputXGradient \citep{shrikumarLearningImportantFeatures2017}, DeepLift \citep{shrikumarLearningImportantFeatures2017}, Guided Backpropagation \citep{springenbergStrivingSimplicityAll2015}, and Integrated Gradients \citep{sundararajanAxiomaticAttributionDeep2017}.
Surrogate models, on the other hand, sample around the input $\x^*$ and use a model's output $f(\x)$ to train a simple, usually linear, model and interpret $f$ through this local approximation.
In this work, we consider the surrogate methods LIME \citep{ribeiroWhyShouldTrust2016} and Kernel SHAP \citep{lundbergUnifiedApproachInterpreting2017}.
Additionally, our study includes Gradient SHAP \citep{lundbergUnifiedApproachInterpreting2017}, an approximation of Shapley value sampling.

We also consider two baselines.
Firstly, we set the explanation for a particular input sequence $\x^*$ to uniformly distributed noise $\s(\x^*) \sim \mathcal{U}(0, 1)^d$. This serves as a null model corresponding to the hypothesis that the XAI method has no knowledge of the informative features $h(\x^*)$. 
Secondly, we employ the Pattern approach \citep{haufeInterpretationWeightVectors2014,wilmingScrutinizingXAIUsing2022}. 
We apply a variant of it by employing the covariance between input features and target $\s_j(\x^*) = \cov(x^*_j, y)$. We call this the Pattern Variant, for which we utilized the tf-idf \citep{sparck1972statistical} representation of each input sequence $\x^{(i)}$.
Clearly, the explanation $\s$ is independent of both the model $f$ and an instance $\x^*$; therefore, it yields the same \revision{feature attributions} for all input sequences.

We apply these XAI methods to all fine-tuning variants of the BERT model and compute explanations on all test data sentences using the default parameters of each method. 
For all XAI methods except LIME, we use their Captum \citep{kokhlikyanCaptumUnifiedGeneric2020} implementation. For LIME, we use the author's original code\footnote{\url{https://github.com/marcotcr/lime}}.

\subsubsection{Explanation Performance Quantification}
For a given instance $\x^* \in \D^{test}$ we aim to quantitatively assess the correctness of its explanation $\s(\x^*)$.
The corresponding ground truth $h(\x^*)$ defines a set of potentially important tokens based on alteration of words; however, a model $f$ might only use a subset of such tokens for its predictions.
Hence, an explanation method that only highlights a subset of tokens that correspond to ground truth, compared to all tokens of the ground truth, must be considered equally correct.
Expressed in information retrieval terms, we are interested in mitigating the impact of false-negatives and emphasizing the impact of false-positives on explanation performance.
False-negatives occur when a token flagged as a ground truth receives a low importance score, and false-positives occur when a feature flagged as not part of the ground truth receives a high importance score.
The Mass Accuracy metric ($\MA$) \cite{arrasCLEVRXAIBenchmarkDataset2022, clarkTetris2023} provides such properties and is defined as
\begin{equation}
    \label{eq:mass-accuracy}
    \MA\left(h(\x^*), \s(\x^*)\right) 
    = \sum_{j=1}^d \s_j(\x^*) h_j(\x^*).
\end{equation}
Here the \revision{feature attributions} $\s$ are normalized, such that $\sum_j^d \s_j = 1$ and  $\s(\x^*) \in [0,1]^d$.
The score $ \MA\left(h(\x^*), \s(\x^*)\right) = 1$ shows a perfect explanation, marking only ground truth tokens as important.
For instance, a sentence with only two ground truth tokens $h(\x^*)=(1,1, 0, 0)^{\top}$, where the attribution for only one ground truth token is high, say $\s(\x^*)=(0.9, 0, 0, 0.1)$, the $\MA$ metric still produces a high score of $ \MA\left(h(\x^*), \s(\x^*)\right) = 0.9$, de-emphasizing false-negatives.
With respect to false-positives, high attributions to non-ground-truth tokens do not directly contribute to the $\MA$, yet, through the normalization of $\s$, all other tokens get assigned a relatively low (non-zero) importance, leading to an overall low $\MA$ score, effectively penalizing false-positive attributions.

\revision{Note, Feature attributions $\s$ are calculated at the sub-word level.
To align them with word-level ground truth, we normalize attribution scores across a sentence and then aggregate sub-word contributions back to the word level.
For example, the word ``benchmark'' may be split into ``bench'' and ``mark'' by the BERT tokenizer, with attributions $\s_{\text{bench}}$ and $\s_{\text{mark}}$, which are combined to $\s_{\text{benchmark}}=\s_{\text{bench}}+\s_{\text{mark}}$.}

\subsubsection{Explanation Bias Quantification}

The change in mass accuracy $\MA$ serves two purposes: (i) It assesses the correctness of XAI methods' output with respect to the ground truth and (ii) with deviations from the ground truth, depending on which layer is fine-tuned or retrained, we can define a notion of what we call \emph{explanation bias}.
Explanation bias is defined via the relative mass accuracy ($\mathrm{RMA}$)
\begin{equation}
    \label{eq:relative-mass-accuracy}
    \mathrm{RMA}_{baseline} \coloneqq \frac{\MA}{\MA_{baseline}},
\end{equation}
where the deviation of explanation performance with regard to a baseline model is quantified, a zero-shot BERT model is used in this work (see next Section \ref{sec:classifiers}).

\subsubsection{Classifiers} \label{sec:classifiers}
In our analysis, we focus on the popular BERT model \citep{devlinBERTPretrainingDeep2019}, though one can expand this work using other common language models such as RoBERTa \citep{liuRoBERTaRobustlyOptimized2019}, XLNet \citep{yangXLNetGeneralizedAutoregressive2019}, or GPT models \citep{radford2018improving,radford2019language}.
For all experiments, we use the pre-trained uncased BERT model \citep{devlinBERTPretrainingDeep2019}\footnote{Hosted by Hugging Face: \url{https://huggingface.co/google-bert/bert-base-uncased}}.
To investigate the impact that fine-tuning or re-training of different parts of BERT's architecture can have on explanation performance, we consider four different training paradigms: (i) We roughly split BERT's architecture into three parts: \textit{Embedding}, \textit{Attention}, and \textit{Classification}. 
The standard approach to adopting BERT for a new downstream task is to train the last classification layer, which we call Classification, while fixing the weights for all remaining parts of the model, here, Embedding and Attention.
We thereby only train a newly initialized classification layer and call the resulting model \textit{BERT}-C. (ii) We additionally train the Embedding layer from scratch, resulting in a model called \textit{BERT}-CE. (iii) In the third model, \textit{BERT}-CEf, the embeddings are fine-tuned as opposed to newly initialized.
In training paradigm (iv), we fine-tune the Embedding and Attention parts of BERT's architecture, resulting in model \textit{BERT}-CEfAf.
Moreover, a zero-shot model \textit{BERT}-ZS, which experienced no gradient updates, was applied.
Lastly, a vanilla one-layer attention model, \textit{OLA}-CEA, comprising a lower-dimensional embedding layer, one attention layer, and a classification layer, was trained from scratch only on the GECO dataset.
\revision{Therefore, without pre-training on external corpora, it represents the simplest attention-based model free from residual biases, providing a clean reference point against which more complex, pre-trained models like BERT can be compared.}
All models achieve an accuracy above or close to 80\% on the test set. Previous works on classification problems involving BERT suggest that accuracy results ranging from $60$ to $90\%$ are standard \citep{8864964,zheng2019new,yuadapting}.
Therefore, we consider our results as evidence that the model has successfully generalized to the given downstream task. 
\revision{Table \ref{tab:bert-models} summarizes model performance with average accuracy and standard deviation over five models trained with different seeds.
More details are given in Appendix \ref{Appendix.Models.Training} and experiments' configuration file\footnote{\url{https://osf.io/74j9s/files/p23yh?view_only=8f80e68d2bba42258da325fa47b9010f}.}}

\section{Experiments and Results}\label{sec:experiments-results}

Our bias analysis on GECO shows that there is no gender bias present in the $\D_A$ dataset with $\bias_C(\D_A^F) = \bias_C(\D_A^M) = \bias_C(\D_A^{NB}) \approx 0.33$.
For dataset $\D_S$ we achieve the scores $\bias_C(\D_S^{NB}) \approx 0.3$, $\bias_C(\D_S^M) \approx 0.36$, and $\bias_C(\D_S^F) \approx 0.33$.
The small difference from a perfect score can be attributed to labeling errors, but is also expected for the dataset $\D_S$ due to its construction, as we only change the human subject of the sentence; other gender terms, referring to other protagonists in the sentence, are kept unchanged.

\begin{table*}[ht]
    \centering
    \begin{tabular}{lccccc}
        \toprule
        \textbf{Models} & \textbf{Embedding} & \textbf{Attention} & \textbf{Classification} & \textbf{Acc. (\%) $\mathcal{D}_A^{test}$} & \textbf{\revision{Acc. (\%)} $\mathcal{D}_S^{test}$} \\
        \midrule
        \textit{BERT}-ZS & fix &  fix & fix & 65.2 $\pm$ 0.0 & 56.6 $\pm$ 0.0 \\
        \textit{BERT}-C & fix &  fix & re-trained & 98.7 $\pm$ 0.1 & 90.2 $\pm$ 0.4 \\
        \textit{BERT}-CE & re-trained & fix & re-trained & 99.8 $\pm$ 0.1 & 96.6 $\pm$ 0.2 \\
        \textit{BERT}-CEf & fine-tuned & fix & re-trained & 99.8 $\pm$ 0.1 & 97.4 $\pm$ 0.3 \\
        \textit{BERT}-CEfAf  & fine-tuned & fine-tuned & re-trained & 99.7 $\pm$ 0.1 & 98.8 $\pm$ 0.2 \\
        \textit{OLA}-CEA & re-trained & re-trained & re-trained & 99.5 $\pm$ 0.2 & 85.1 $\pm$ 14.6 \\
        \bottomrule \\
    \end{tabular}
        \caption{Overview of BERT transfer learning paradigms and the performance of the resulting models on the test datasets $\D_A^{test}$ and $\D_S^{test}$. 
        }
    \label{tab:bert-models}
\end{table*}

\begin{figure}
    \centering
    \begin{subfigure}[b]{0.89\textwidth}
        \includegraphics[width=\linewidth]{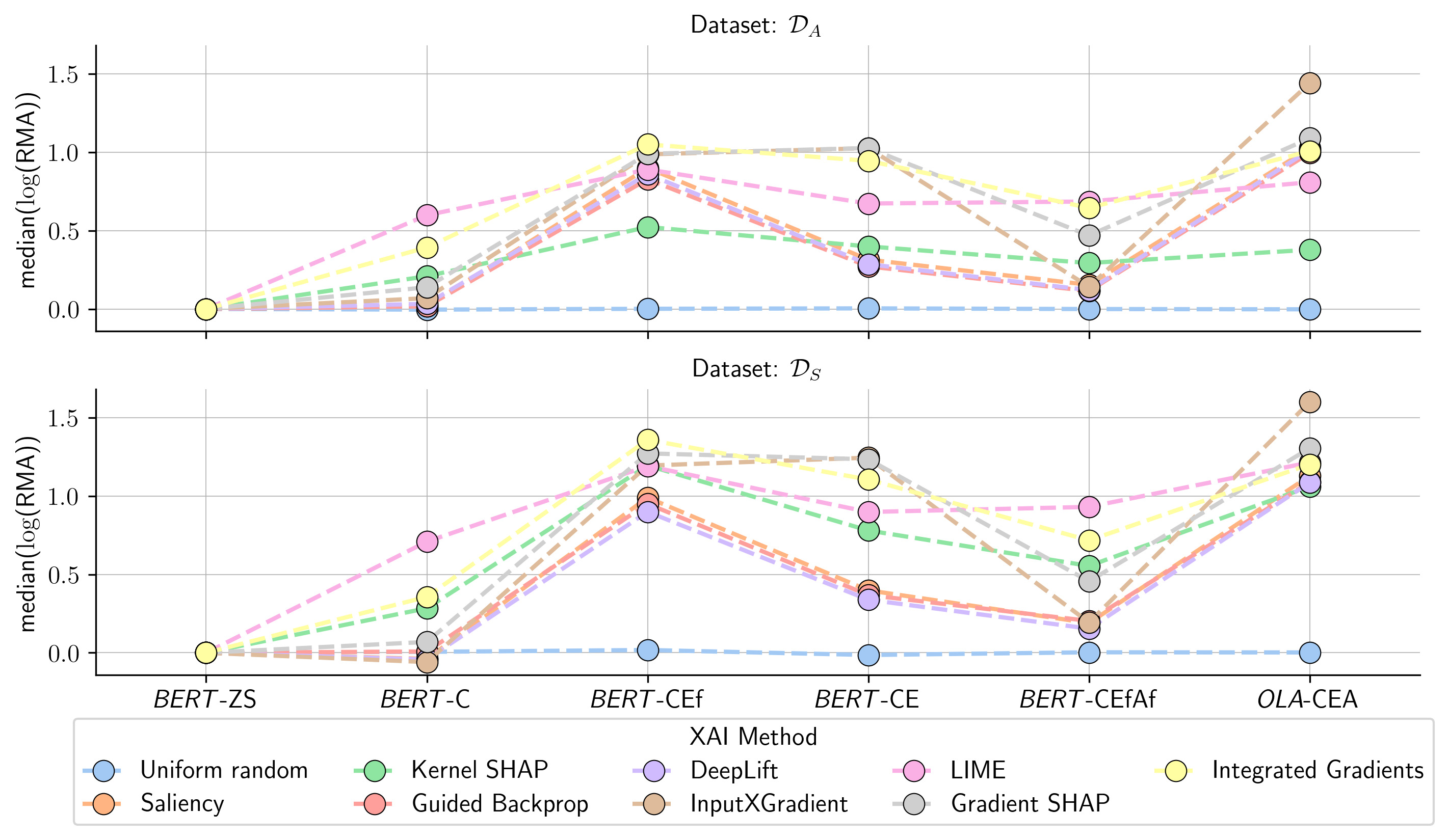}
        \caption{}
        \label{fig:explanation-performance-gender-rma}
    \end{subfigure}  
   
    \begin{subfigure}[b]{0.89\textwidth}
        \includegraphics[width=\linewidth]{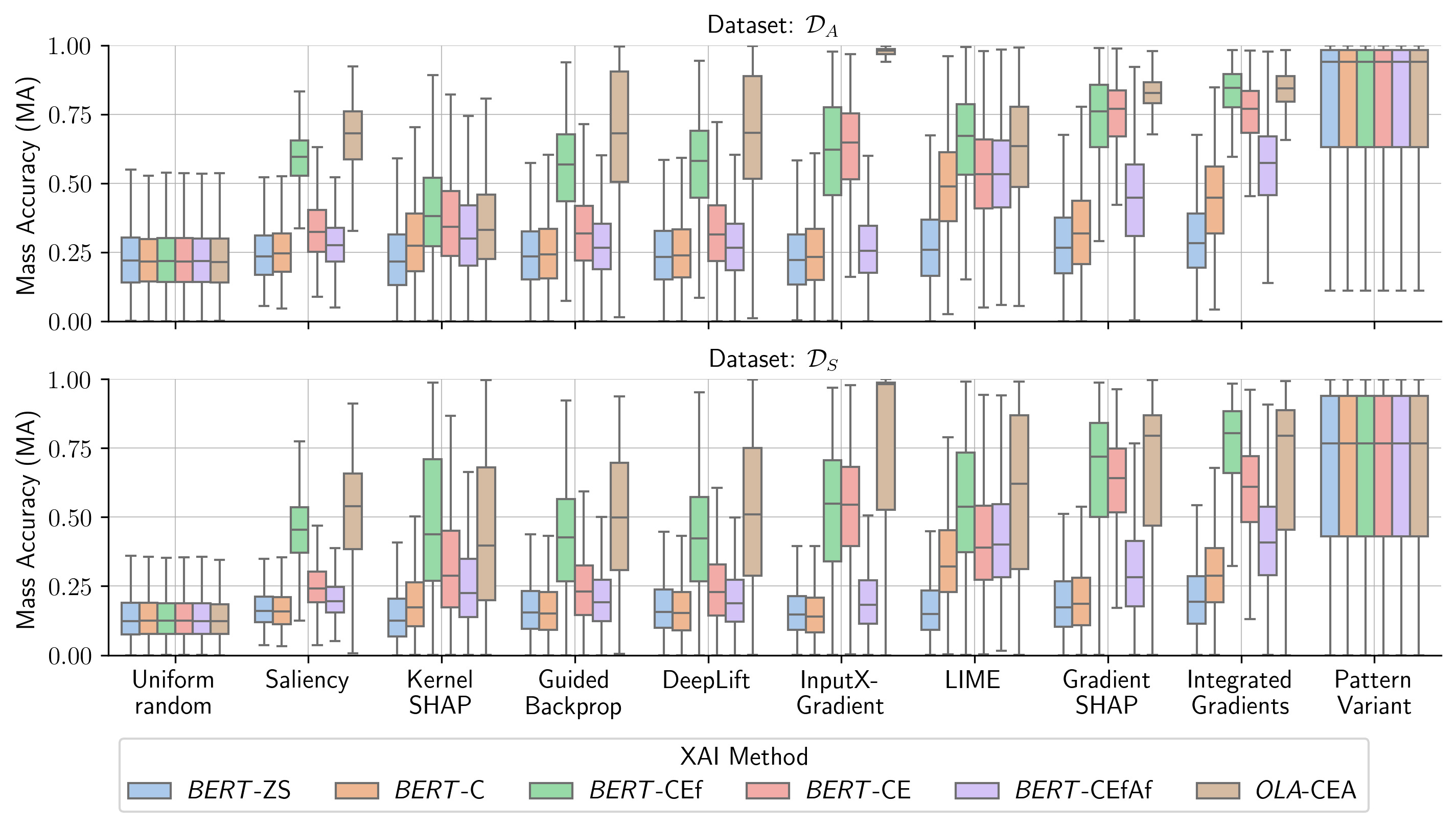}
        \caption{}
        \label{fig:explanation-performance-gender-ma}
    \end{subfigure}

    \caption{Explanation performance of different post-hoc XAI methods applied to language models that were adapted from BERT using five different transfer learning schemes. 
    XAI evaluations were carried out on classified sentences in two gender-classification tasks, represented by datasets $\D_S$ and $\D_A$. 
    The baseline performance for uniformly drawn random \revision{feature attributions} is denoted by \textit{Uniform Random}. \textit{Pattern Variant} denotes a model- and pretraining-agnostic global explanation method. In (a), the relative change in explanation performance with respect to a zero-shot BERT model shows consistent changes for models with fine-tuned embeddings.
    In (b), fine-tuning or retraining of the embedding layers of \textit{BERT} leads to consistent improvements in explanation correctness even when model performance is held constant for all models. 
    Applying XAI methods to the \textit{OLA} model leads to overall higher explanation performance, with InputXGradient becoming on par with Pattern Variant.}
    \label{fig:explanation-performance-gender}
\end{figure}

Using the unbiased dataset GECO, we conduct experiments to study the influence of biased models on explanation performance.
After fine-tuning and re-training the models (see Table \ref{tab:bert-models}), we apply \revision{feature attribution} methods.
Figure \ref{fig:explanation-performance-gender} shows the explanation performance of sample-based \revision{attribution} maps $\s(\x^{*})$ produced by the selected \revision{feature attribution} and baseline methods. 

\revision{In the following, we present the results towards the research questions RQ1 and RQ2.
Firstly, we present the results for explanation performance, addressing questions about the data-centric correctness of the analyzed feature attribution methods.
Secondly, we present bias-related results, focusing on how the correctness of feature attribution indicates biases in language models.}

\textbf{Regarding RQ1:} We observe a general difference in $\MA$ between datasets $\D_A$ and $\D_{S}$. 
While for the majority of \revision{attribution} methods, the performance for experiments on dataset $\D_{S}$ stays on a level lower than $0.25$, experiments on dataset $\D_A$ are often able to offset these results into levels above $0.25$ (see Figure \ref{fig:explanation-performance-gender-ma}).
However, dataset $\D_{S}$ has fewer altered gender words, thus fewer discriminative tokens, leading to an overall degradation of classification accuracy across all models (see Table \ref{tab:bert-models}), which also impacts explanation performance.
For all BERT models and both datasets, Integrated Gradients consistently outperforms other methods and the Uniform random baseline.
\revision{Results for both $\D_{A}$ and $\D_{S}$ show that LIME, Gradient SHAP, and Integrated Gradient are the highest-performing methods for all BERT models, compared to the Pattern Variant baseline, with respect to the data-centric SAP criterion (see Definition \ref{def:sap}).}

\revision{
Comparing the \textit{OLA}-CEA model to all BERT models, we observe a stark contrast in explanation performance. 
Recall that the \textit{OLA}-CEA model was purely trained from scratch on the gender-controlled dataset GECO; hence, it does not suffer from any gender bias.
The mass accuracy between both datasets is similar, with higher variance for some methods, such as Guided Backprop and DeepLift, yet the overall performance of almost all explanation methods significantly increases compared to the random baseline.
In addition to relatively well-performing methods like Integrated Gradients, LIME, and Gradient SHAP, the $\MA$ of InputXGradient comes very close to the covariance baseline, making it the best-performing method.}

\revision{Although no explanation method achieves the correctness score of Pattern Variant, fine-tuning a biased embedding layer for a downstream task has a high impact on the output for some methods.
The Pattern Variant is a model-independent global explanation method that relies solely on the intrinsic structure of the data itself. 
It performs optimally when the feature and target relation is governed by a linear relationship, which is mainly the case for GECO.
This can be seen in Figure \ref{fig:app:correlation-plot-data} of the appendix, where we visualize the Pearson correlation of the term frequency–inverse document frequency (tf-idf) \citep{sparck1972statistical} representation of words and the target, clearly showing how we infuse dependency through the word alteration procedure.}

\revision{Exemplarily, Figure \ref{fig:sample-explanation-gender} in the appendix highlights a sentence labeled as ``female'', which is shown together with its word-based \revision{feature attributions} as bar plots for each fine-tuning stage.
We observe a high variability in token attribution between differently fine-tuned BERT models, and the pronoun ``she'' receives relatively high importance compared to other words.
However, not all XAI methods agree on the importance of the token ``she''; for example, for model \textit{BERT}-CE InputXGradient attributes high importance to it, yet for model \textit{BERT}-CEfAf, it attributes rather high importance to the word ``Bella.''}

\textbf{Regarding RQ2:} In Figure \ref{fig:explanation-performance-gender-rma} we can observe a consistent pattern with respect to the fine-tuning stages across both data scenarios $\D_{S}$ and $\D_A$ in terms of $\mathrm{RMA}_{BERT-\text{ZS}}$ performance.
\revision{Here, \textit{BERT}-ZS is utilized as a baseline model, as it represents the ``untouched'' pre-trained model without gradient updates. 
Using it as a baseline allows us to quantify how the performance of feature attribution correctness evolves as models become increasingly specialized in the gender-classification task.
Specifically, it highlights how fine-tuned models focus more on the discriminatory tokens compared to the zero-shot model, thereby providing a relative quantification of the mitigation of residual bias.}
For the scenario $\D_A$, it is clear that the models \textit{BERT}-CE and \textit{BERT}-CEf, where the embedding layer was trained or fine-tuned, respectively, outperform \textit{BERT}-C and \textit{BERT}-CEfAf (see also Figure \ref{fig:explanation-performance-gender-ma}).
This shows that the embeddings encode numerous bias information and, indeed, influence \revision{data-centric} explanation performance.

\section{Discussion}

With GECO and GECOBench, we propose an open framework for benchmarking the correctness of \revision{feature attributions} of pre-trained language models as well as aspects of fairness. 
Our initial results demonstrate (a) differences in explanation performance between \revision{feature attribution} methods, (b) a general dependency of explanation performance on the amount of re-training/fine-tuning of BERT models, and (c) residual gender biases as contributors to sub-par explanation performance. 

More generally, the proposed gender classification problem is a simplification that does not reflect the complexity and diversity of gender identification in our world today; however, by providing non-binary gendered sentences, we attempt to counteract historical gender norms and provide a more inclusive basis for gender-bias research.
We view the gender classification task as a minimal proxy for the gender bias issue, modeling all necessary properties to analyze bias propagation into \revision{feature attribution methods}. 
We also view predicting a sentence's gender as an auxiliary task, which we consider more of an academic problem that naturally arises from how we construct sentences, but has, as we see it, no immediate application or societal impact.
\revision{While GECO provides a controlled environment to study how gender bias influences feature attributions, we acknowledge that our design inevitably oversimplifies gender by restricting it to pronoun alterations (e.g., he/she/they). 
This simplification risks reinforcing notions of gender that fail to represent the full spectrum of identities.
In addition, gender-controlled datasets such as GECO could be misused, for example, to build or evaluate models explicitly aimed at gender classification rather than for bias analysis. 
To mitigate this risk, we emphasize that GECO is intended solely for studying the correctness assessment of feature attribution methods under controlled bias conditions, not for downstream applications involving sensitive demographic prediction. 
We view GECO as a first step toward systematic evaluation of bias in feature attributions, with the expectation that future work will extend its coverage to more diverse text sources, richer notions of gender, and broader fairness concepts.}
\revision{While direct extrapolation to more complex applications is challenging, our results indicate that feature attributions are indeed affected by gender bias. 
This motivates caution in downstream tasks such as sentiment analysis or toxic language detection, where attribution methods might wrongly highlight some gender-related tokens, due to bias rather than semantic relevance. 
For example, this renders model debugging tasks challenging, as developers and researchers cannot distinguish between feature attributions that suggest ``flaws'' in the model, arising from genuine model bias or from other artifacts of the data.}

When it comes to data selection, we are convinced that if biased models are applied to semantically gender-neutral sentences, no reliance on words representing classical gender roles can be expected; thus, the potential impact of biases on \revision{feature attributions} cannot be measured.
Thus, the sentences selected to create GECO were intentionally taken from Wikipedia articles outlining the storylines of classic novels due to their likely use of historical gender norms. 
These are, indeed, prerequisites for assessing the behavior of XAI methods’ output applied to language models experiencing various levels of biases. 

By creating grammatical female, male, and non-binary versions of a particular sentence based on pronouns, we are aiming to break such historical gender associations with respect to the \emph{classification task} represented by the datasets $\D_S$ and $\D_A$.
Single sentences may still entail historical gender associations, which can be utilized by biased machine learning models. 
However, the classification task arising from GECO is gender-balanced and models specialized in that task, through successive fine-tuning and re-training of an increasing number of layers, learn to not rely on such historical gender associations, as altered pronouns are, by construction, the only words associated with the prediction target. 
It can then be shown that models basing their decisions on words other than the words altered by us have learned stereotypical associations.
For example, consider the sentence ``She prepares dinner in the kitchen while he is outside fixing the car.''
This sentence illustrates ``traditional'' gender roles, where the woman is associated with domestic tasks and the man with mechanical or manual labor, reinforcing stereotypes. 
By altering the pronouns such that the sentence only contains ``she'' or ``he'' pronouns, we ``break'' these stereotypical gender roles. 
However, only partially in this instance will one part of the sentence always reflect traditional gender roles, e.g.: ``She prepares dinner in the kitchen... .'' 
Yet, for the classification task represented by the datasets $\D_S$ and $\D_A$, only the altered \revision{words} represent a relationship with the prediction target.
Biased language models might then leverage words like ``kitchen'' or ``car'' for their decision, and unbiased models must only rely on \revision{altered words}, and historical gender norms embedded in sentences become irrelevant.
\revision{In Future research, co-reference resolution could be an immediate extension of GECO because it takes the same gender-manipulated sentences and asks the model not just to classify gender per sentence, but to resolve references consistently across discourse, thereby testing explanation correctness under contextual and bias-sensitive conditions}

\revision{In terms of data-centric correctness assessments of feature attribution methods, Pattern Variant indeed offers strong theoretical justification for detecting important features according to statistical associations \citep{haufeInterpretationWeightVectors2014}, establishing a solid baseline for the upper bound of explanation performance in our benchmark.}
Compared to the random baseline, we observe two further high-performing \revision{attribution} methods in the transfer learning regime \revision{(in terms of SAP (see Definition \ref{def:sap}))}: Integrated Gradients and Gradient SHAP. 
Yet, these methods still do not reach the mass accuracy level of Pattern Variant. 
The reasons can be two-fold: (i) As shown by \citep{clarkTetris2023} and \citep{wilmingTheory2023}, \revision{feature attribution} methods consistently attribute importance to suppressor variables, features not statistically associated with the target but utilized by machine learning models to increase accuracy.
And (ii), model bias impacts \revision{feature attributions}. 
\revision{We show that the gender bias in BERT leads to residual asymmetries in feature attributions and forms a consistent pattern of deviation in correctness, depending on which layer of BERT was fine-tuned or re-trained, while still achieving equivalent classification accuracy.}
As a result, updating embedding layers has the strongest impact on \revision{feature attributions}.
\revision{These findings indicate that embeddings contain significant bias affecting feature attribution methods, and that the proposed data-centric notion of correctness of feature importance is indicative of model bias.}

While this is, to our knowledge, the first XAI benchmark addressing a well-defined notion of \revision{data-centric correctness of feature importance} in the NLP domain, we do not consider it an exhaustive evaluation of \revision{feature attribution} methods but rather a first step towards this. 
A possible limitation of our approach is that the criterion of univariate statistical association used here to define important features or tokens does not include nonlinear feature interactions that are present in many real-world applications. 
However, for analyzing the fundamental behaviors of \revision{feature attribution} methods, this characteristic allows for straightforward evaluation strategies, permitting us to embed these statistical properties into the proposed corpus and establish a ground truth of word relevance.
Designing metrics for evaluating explanation performance -- especially for measuring correctness -- is another subject that requires further research.

\section{Conclusion}

We have introduced GECO -- a novel gender-controlled ground truth text dataset designed for the development and evaluation of \revision{feature attribution} methods -- and GECOBench -- a quantitative benchmarking framework to perform objective assessments of explanation performance for language models.
We demonstrated the use of GECO and GECOBench by applying them to the pre-trained language model BERT, a model known to exhibit gender biases. 
With this analysis, we showed that \revision{the SAP criterion is an effective condition to quantify the data-centric correctness of feature attribution methods applied to the language model BERT, and that residual biases contained in BERT affect feature attributions and can be mitigated through fine-tuning and re-training of different layers of BERT, positively impacting explanation performance}.

\section*{Conflict of Interest Statement}
The authors declare that the research was conducted in the absence of any commercial or financial relationships that could be construed as a potential conflict of interest.

\section*{Author Contributions}
RW: Formal analysis, Methodology, Software, Validation, Writing – original draft, Writing – review \& editing, Conceptualization, Data curation, Visualization. AD: Formal analysis, Methodology, Software, Validation, Writing – original draft, Writing – review \& editing, Data curation, Visualization. HS: Formal analysis, Methodology, Software, Validation, Writing – original draft, Writing – review \& editing, Data curation, Visualization. MO: Software, Validation, Conceptualization, Data curation. BC: Formal analysis, Methodology, Software, Validation, Writing – review \& editing, Data curation. SH: Conceptualization, Funding acquisition, Investigation, Supervision, Writing – original draft, Writing – review \& editing, Methodology, Project administration, Resources.

\section*{Funding}
This result is part of a project that has received funding from the European Research Council (ERC) under the European Union’s Horizon 2020 research and innovation programme (Grant agreement No. 758985), the German Federal Ministry for Economy and Climate Action (BMWK) in the frame of the QI-Digital Initiative.

\bibliographystyle{icml2025}
\bibliography{references-all}

\appendix

\section{Appendix}
\subsection{GECO Dataset}\label{Appendix.Gender_Dataset}

\subsubsection{Data Licensing}\label{app:data-licensing}
The titles from Project Gutenberg are available under a public domain license, allowing them to be freely accessed and used by the public.
However, our datasets are primarily based on Wikipedia articles. Wikipedia articles are available under the Creative Commons Attribution-ShareAlike 4.0 International License, allowing us to remix, transform, and redistribute the material.

In the following, we describe the data generation and format in more detail and perform a bias analysis as a sanity check to verify that the dataset is unbiased.

\subsubsection{Data Generation}\label{Appendix.Gender_Dataset.Data_Generation}
We accessed the top 100 list of popular books on Project Gutenberg on March 17, 2022, and to obtain the corresponding Wikipedia articles, we ran Google queries.
For web scraping, we use the software Selenium\footnote{https://www.selenium.dev/}. After scraping the sentences from the Wikipedia articles of the books, we preprocess the sentences using the Python library Spacy\footnote{https://spacy.io}. 

Below, we provide additional details of our data processing rules as part of the data generation process. We employ Spacy to only include sentences with root verbs in the 3rd person singular.

\revision{The objective of this dataset is to construct a ground truth resource for evaluating language models under controlled grammatical-gender manipulations. 
To this end, we created a set of manipulated sentences, each available in three variants: male, female, and non-binary. Depending on the manipulation scheme, sentences either (i) replace only the subject’s gendered words or (ii) replace all gender-related words (subjects, objects, and modifiers) to produce fully male or fully female versions.
The inclusion of non-binary variants was intended to provide a more inclusive basis for bias research, though we acknowledge that the representation of gender through pronoun alternation remains a simplification.}

\revision{\paragraph{Sentence segmentation and filtering}}
\revision{We segmented over 5000 paragraphs into candidate sentences using the small English model of Spacy, yielding more than 19,500 sentences. 
Spacy’s syntactic annotations were then leveraged to filter the sentences with:
(i) root verb selection: we kept only sentences whose root verb was in third-person singular, determined from Spacy’s morph feature annotations.
(ii) subject extraction: subjects were identified using the dependency relation nsubj (or closely related subject arcs) relative to the root verb. 
We converted Spacy’s dependency generator to a list for each candidate sentence.
(iii) sentence length constraint: to improve clarity and reduce annotation burden, we discarded sentences with more than 30 tokens. 
(iv) punctuation filter: sentences not ending with a period were removed.
(v) exclusion of ``it'' subjects: we filtered out sentences where the subject was ``it'', as these do not contribute to gender-based evaluation.}

\revision{Identification of human-related subjects}
\revision{To restrict the dataset to human-relevant entities, we retained only subjects categorized as:
(i) proper nouns (Spacy: token.pos\_ == ``PROPN'') referring to characters.
Author names and spurious references (e.g., scholars writing about the book) were manually removed.
(ii) pronouns (``he''/``she''), detected by Spacy’s morph features: PronType=Prs and $\text{Gender} \in \{\text{Masc},\text{Fem}\}$.
(iii) common nouns (e.g., father, sister, boy, girl), which were manually screened to ensure they referred to human roles rather than misclassifications or irrelevant objects.
To identify human-related nouns, we compiled lists for each group (proper nouns, pronouns, common nouns) and manually validated them. 
Single-occurrence names were typically excluded, as they often referred to non-plot entities.}

\revision{\paragraph{De-duplication and manual review}}
\revision{We removed duplicate sentences and further manually inspected candidate sentences with a custom visualization tool, highlighting the detected subject and root verb. 
Sentences that were malformed, fragmentary, or unrelated to the plot (e.g., citations, chapter headings, or errors in Spacy segmentation) were excluded.}

\revision{\paragraph{Sentence annotation and gender manipulation}}
\revision{
From this filtering process, we obtained 4,830 sentences.
Each was then manually annotated using a custom-built web interface\footnote{\url{https://github.com/braindatalab/gecobench/tree/main/data/dataset_generation}}.
Annotation was conducted by four annotators, with disagreements resolved through discussion. 
Inter-annotator agreement was not formally measured (given the deterministic nature of replacements), but the four annotators reviewed every sentence, and residual errors were corrected.}

\revision{For each sentence, we generated three variants:
(i) male version: subjects, pronouns, and gendered words were replaced with male forms (e.g., she $\to$ he, daughter $\to$ son).
(ii) female version: parallel substitutions for female forms.
(iii) non-binary version: we used the singular they pronoun, adjusting possessives (her $\to$ their) and verbs for agreement (she thinks $\to$ they think).
Proper names were replaced with the appropriate pronoun (e.g., Paul did $\to$ He did, She did, They did). 
Gender-indicative common nouns (e.g., son, brother, king) were mapped to gendered counterparts, with neutral forms chosen where available (sibling, monarch, child). 
Special care was taken to ensure that verb agreement remained grammatical in the non-binary variants.}

\revision{
This pipeline resulted in two ground-truth datasets:
$\mathcal{D}_S$: only subject terms are manipulated.
$\mathcal{D}_A$: all gendered terms are manipulated.
}
\revision{
Together, these datasets provide 3 variants of 1,610 base sentences (9,660 total), each aligned at the token level and accompanied by word-level labels for ground-truth feature attributions.
}
\begin{table}[t]
\centering
\caption{\revision{Labeling rules for the two dataset variants. $\mathcal{D}_S$ manipulates only the subject term, while $\mathcal{D}_A$ manipulates all gender-related words.}}
\label{tab:labeling-rules}
\begin{tabular}{p{7cm} p{7cm}}
\toprule
\textbf{$\mathcal{D}_S$: Subject-only manipulations} & \textbf{$\mathcal{D}_A$: All gender-related manipulations} \\
\midrule
\begin{itemize}
  \item Replace only the grammatical subject (proper nouns, pronouns, or human-related common nouns).
  \item Proper names $\to$ pronouns (e.g., \emph{Paul did} $\to$ \emph{He/She/They did}).
  \item Pronoun subjects: \emph{he/she/they}.
  \item Subject common nouns: \emph{son $\to$ daughter $\to$ child}.
  \item Verb agreement corrected for non-binary forms (\emph{She thinks $\to$ They think}).
  \item All other tokens remain unchanged.
\end{itemize}
&
\begin{itemize}
  \item Replace all gendered tokens: subjects, objects, modifiers, kinship terms.
  \item Apply subject rules as in $\mathcal{D}_S$.
  \item Objects: \emph{her brother $\to$ his sister $\to$ their sibling}.
  \item Possessives: \emph{his/her/their}.
  \item Kinship/titles: \emph{king $\to$ queen $\to$ monarch}, \emph{father $\to$ mother $\to$ parent}.
  \item Ensure correct verb agreement with singular \emph{they}.
  \item Sentences systematically checked for grammatical well-formedness, otherwise removed.
\end{itemize}
\\
\bottomrule
\end{tabular}
\end{table}

%



We attempted to employ fully automated sentence labeling using GPT-4 \cite{gpt4}, but encountered inconsistencies in identifying names, genders, and gendered terms, as well as detecting human subjects, particularly in dataset $\D_S$. Due to the need for precise ground truth labels to benchmark various explanation methods, we opted for a manual labeling approach instead.

\subsubsection{Data Format}
\label{Appendix.Gender_Dataset.Data_Format}

The datasets are available in the following folder structure. \\

\dirtree{%
    .1 GECO. 
    .2 data\_config.json. 
    .2 gender\_all. 
    .3 test.jsonl (644 Sentences). 
    .3 train.jsonl (2576 Sentences). 
    .2 gender\_subj. 
    .3 test.jsonl (644 Sentences). 
    .3 train.jsonl (2576 Sentences). 
}

The sentences are available in JSONL files, with each line representing a sentence in either male or female form.
An example sentence can be seen in Listing \ref{app:lst:format}.
Each line contains the input sentence as a list of words, as well as the explanation ground truth for each word. The ``target'' field indicates whether the sentence is in female form (0) or male form (1). Lastly, the ``sentence\_idx'' field identifies which original sentence was altered and can be used to match the male and female forms of a sentence. \\

\begin{minipage}{\linewidth}

\begin{lstlisting}[language=Java, caption=Example sentence of the class $\x_A^M$., label=app:lst:format]
{
  "sentence": ["Paul", "loves", "his", "dog"],
  "ground_truth": [1.0, 0.0, 1.0, 0.0],
  "target": 1,
  "sentence_idx": 0
}
\end{lstlisting}
\end{minipage}

\begin{figure*}[ht]
    \centering
    \includegraphics[width=0.9\linewidth]{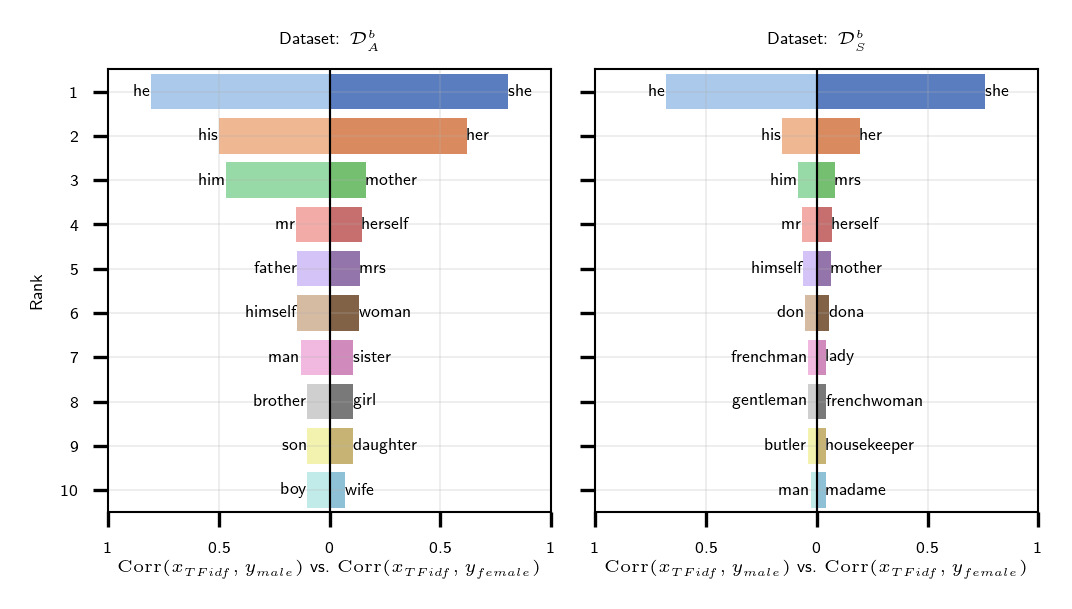}
    \caption{Pearson correlation between tf-idf representation of words and the target for GECO. Here, we see the top ten words by correlation, and labeled words such as the pronouns \textit{he} and \textit{she} or \textit{his} and \textit{her} are consistently ranked highest in both datasets $\mathcal{D}_{A}$ and $\mathcal{D}_{S}$, indicating how the labeling introduced dependency between target and words.}
    \label{fig:app:correlation-plot-data}
\end{figure*}

\subsection{Models}

\revision{In the following, we provide further details about the model training and performance.} All models are implemented in PyTorch. 

\subsubsection{Training}
\label{Appendix.Models.Training}
For model training, we split the two datasets $\D_{S}$ and $\D_A$ into training and test sets $\D_{S}^{train}, \D_{S}^{test}$ and $\D_{A}^{train}, \D_A^{test}$.
\revision{We train $5$ repetitions for each model, using a different random seed for each repetition.}
This is done not only to compensate for variations in model performance \cite{dodge2020fine,bugliarelloMultimodalPretrainingUnmasked2021} but also to capture the resulting variance in model explanations.

We optimize the learning rate and keep the remaining hyperparameters fixed. Table \ref{app:tab:bert-model-hyperparams} shows a full overview of all hyperparameters and values we use. 
While it could be interesting to explore how various training parameters impact explanation performance, our focus is on achieving the same accuracy threshold across all models. 
Therefore, we do not use the same hyperparameters for all models. All models were trained on our internal cluster with an Nvidia A100 (40GB) GPU. 
\revision{For hyperparameter optimization and the final model training across the five different training schemes and two datasets, we conducted 464 training runs with an average running time of 68 seconds.}

\begin{figure*}[ht]
    \centering
    \includegraphics[width=0.9\linewidth]{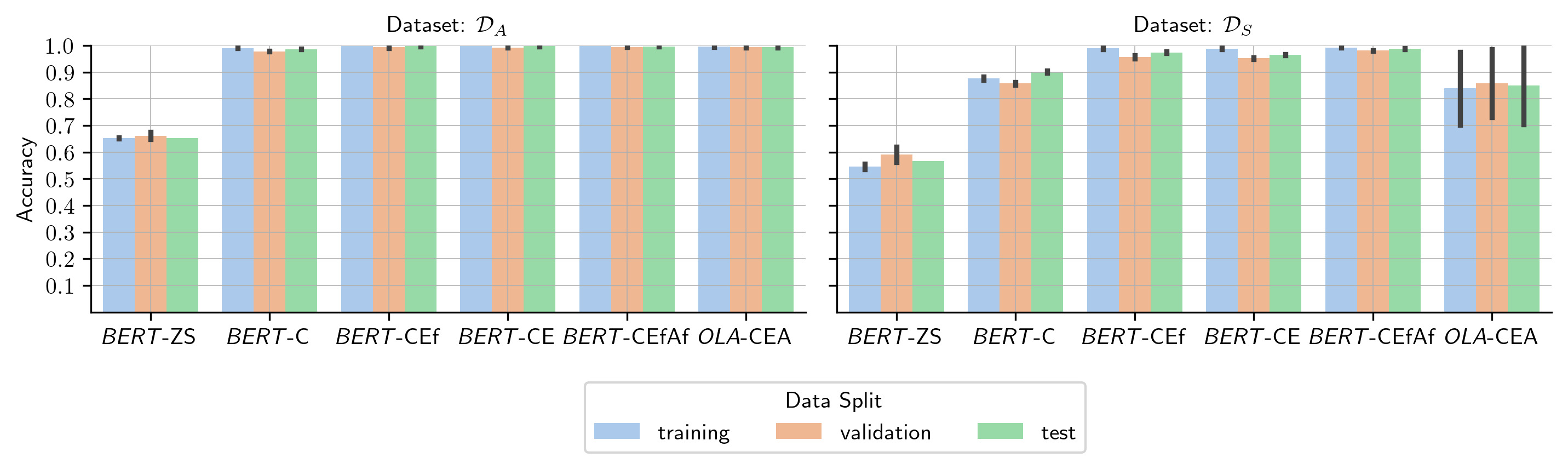}
    \caption{Model performance.}
    \label{fig:app:model_performance}
\end{figure*}

An overview of the performance of the models on the training, validation, and test sets is shown in Figure \ref{fig:app:model_performance}.

\begin{table*}[htbp]
    \centering
    \begin{tabular}{lccccc}
        \toprule
        \textbf{Models} & Batch Size & Embedding Dimension & Epochs & Learning Rate \\
        \midrule
        \textit{BERT}-C & 32 & 768 & 20 & 0.01\\
        \textit{BERT}-CE & 32 & 768 & 20 & 0.0001\\
        \textit{BERT}-CEf & 32 & 768 & 20 & 0.01\\
        \textit{BERT}-CEfAf  & 32 & 768 & 20 & 0.000005 \\
        \textit{OLA}-CEA & 64 & 64 & 200 & 0.01 \\
        \bottomrule \\
    \end{tabular}
    \caption{Overview of BERT training schemes and values of the hyperparameters used to train them.}
    \label{app:tab:bert-model-hyperparams}
\end{table*}

\subsection{Explanations}\label{app.explanations}

\begin{figure}[h!]
     \centering
     \includegraphics[width=0.88\textwidth]{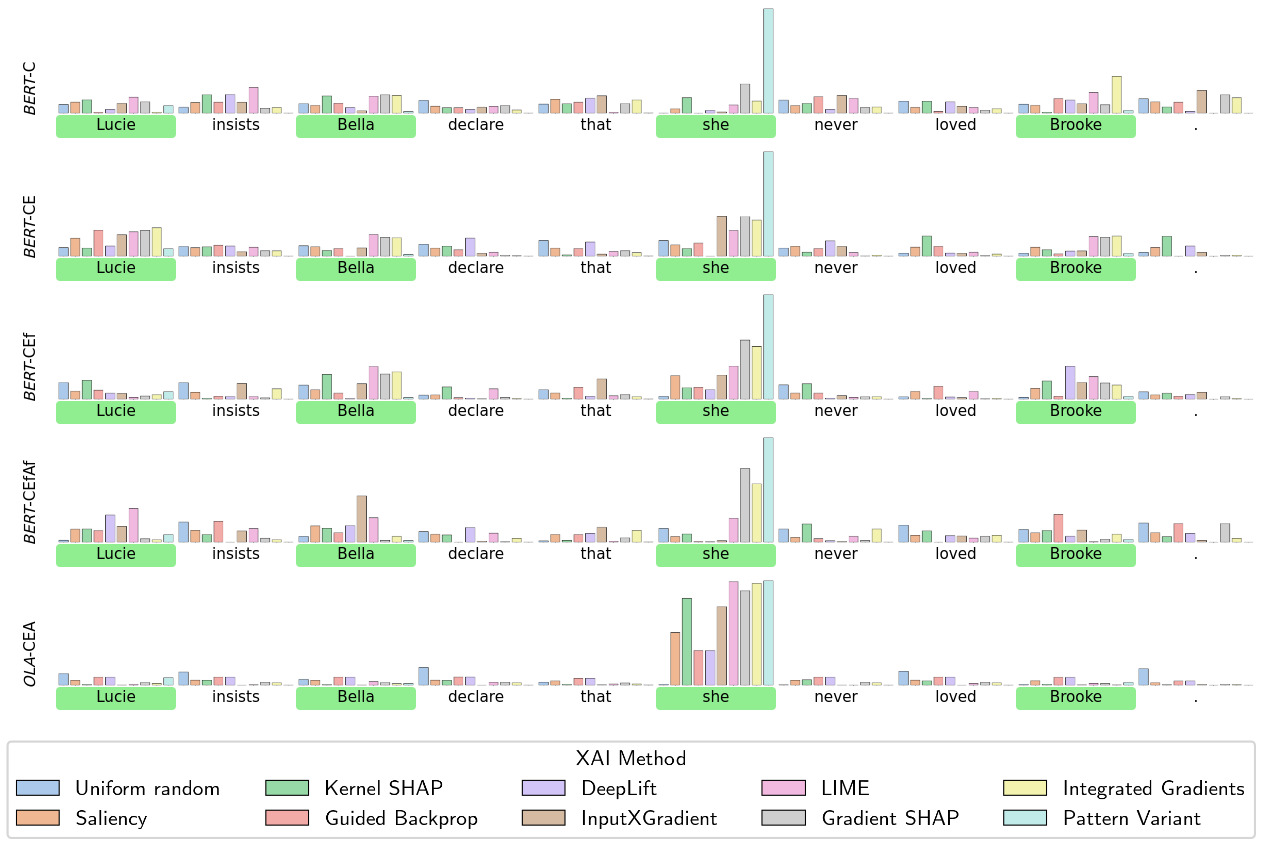}
     \caption{\revision{Feature attributions} by popular explanation methods for one sample sentence, broken down into input tokens as given to the respective model, with the ground truth manipulations highlighted in green.
     The majority of importance by many methods is correctly attributed to the word ``she''; however, all tokenized words show non-zero attribution for multiple methods, including the character period ``.''.}
     \label{fig:sample-explanation-gender}
\end{figure}

\end{document}